\documentclass[10pt,conference]{IEEEtran}
\IEEEoverridecommandlockouts
\usepackage{cite}
\usepackage{amsmath,amssymb,amsfonts}
\usepackage{algorithmic}
\usepackage{graphicx}
\usepackage{textcomp}
\usepackage{xcolor}
\usepackage{caption}
\usepackage{tikz}
\usetikzlibrary{positioning, arrows.meta, calc}
\usepackage{booktabs} 
\usepackage{threeparttable}
\usepackage{gensymb}
\usepackage{float}
\usepackage[numbers, square]{natbib} 
\usepackage{hyperref}
\usepackage{stfloats}
\usepackage{subcaption}

\def\BibTeX{{\rm B\kern-.05em{\sc i\kern-.025em b}\kern-.08em
    T\kern-.1667em\lower.7ex\hbox{E}\kern-.125emX}}
\begin{document}

\title{Hybrid Deep Learning for Traceability and Classification of Industrial Slate Tiles}
\author{
     \IEEEauthorblockN{
         Soren Antebi\textsuperscript{1}, 
         Stefan Eickeler\textsuperscript{1}, 
         Sandra Halscheidt\textsuperscript{1}, 
         René Schmitz\textsuperscript{2}, 
         Michael Müllers\textsuperscript{2}, 
         Dirk Hecker\textsuperscript{1}, 
         and Rafet Sifa\textsuperscript{1}
     }
     \IEEEauthorblockA{\textsuperscript{1}Fraunhofer Institute IAIS, Sankt Augustin, Germany}
     \IEEEauthorblockA{\textsuperscript{2}Rathscheck Schiefer GmbH, Mayen, Germany}
 }
\maketitle
\begin{abstract}
Applying deep learning to instance-aware re-identification of slate tiles and extraction site classification can improve production efficiency and quality control in the slate tile industry. These tasks are particularly important for handling natural materials where visual variability can make manual inspection costly and error-prone. We present a lightweight, hybrid deep learning approach that combines image matching and classification within a single framework. The system integrates a feature-matching branch based on XFeat with a MobileNetV3-based classification branch. The XFeat branch, combined with a LightGlue matching head, improves instance matching performance by +15.4\% AUC. For classification, features from both backbones are shared and fused, resulting in a +10.9\% accuracy improvement over a standard MobileNetV3 model. Our approach is evaluated on a newly created industrial dataset consisting of 2,610 slate tile images from six extraction sites. The results demonstrate the effectiveness of the proposed approach for object re-identification and classification in an industrial setting.
\end{abstract}

\section{Introduction}

Natural slate tiles are widely used in roofing and facade applications due to their durability, environmental sustainability, and longevity \cite{cotneyUnderstandingBenefitsNatural2022}. Slate is extracted from diverse geological regions and processed into tiles with distinct variations in color and texture. The quality of processed slate is commonly assessed based on visual and physical properties \cite{C406C406M15StandardSpecification}, which traditionally, has relied on skilled experts who classify material conditions using tacit knowledge. However, the difficulty of transferring this specialized knowledge means that its gradual loss creates challenges for future quality assurance and ensuring reliable traceability \cite{fenoglioTacitKnowledgeElicitation2022}\cite{johnsonHowWhyWe2019}.

Accurate classification of slate quality and quarry origin is essential for both production and distribution, as the longevity of a tile strongly depends on its material integrity. For example, low-quality slate containing internal fractures is more susceptible to weathering and mechanical stress, increasing the risk of breakage or flaking. Furthermore, determining the quarry of origin is important when specific colors or structural characteristics are required for architectural consistency. From a logistical perspective, reliable re-identification of tiles is critical during handling and return processes, enabling manufacturers to trace individual items. Therefore, classification and re-identification is vital for consistent tracking across production and automated quality assurance pipelines.

Models can help address these challenges by partially replacing or complementing expert-based inspection. However, developing reliable automated systems for slate quality assessment remains challenging due to the subtle variations in surface texture and morphology. For instance, tiles originating from different quarries may be geologically and visually similar. As traditional computer vision approaches might struggle with this aspect, we suggest implementing hybrid deep learning methods that combine locally sensitive and global feature representations \cite{liSlateDetectionOrthophotos2025}\cite{karimiDeepLearningbasedAutomated2024}.

In practical industrial settings, computational constraints further influence system design. High-accuracy deep learning methods frequently rely on computationally intensive architectures \cite{edstedtRoMaRobustDense2023}\cite{ sarlinSuperGlueLearningFeature2020}\cite{ heDeepResidualLearning2015}, which limits their applicability on edge devices and embedded platforms. In production line processes and real-time quality control, computational efficiency and low energy consumption is important. Therefore, there is a strong need for lightweight yet robust models that balance accuracy and efficiency for deployment in resource-constrained environments.
\begin{figure}[t]
\centering

\includegraphics[width=0.9\linewidth]{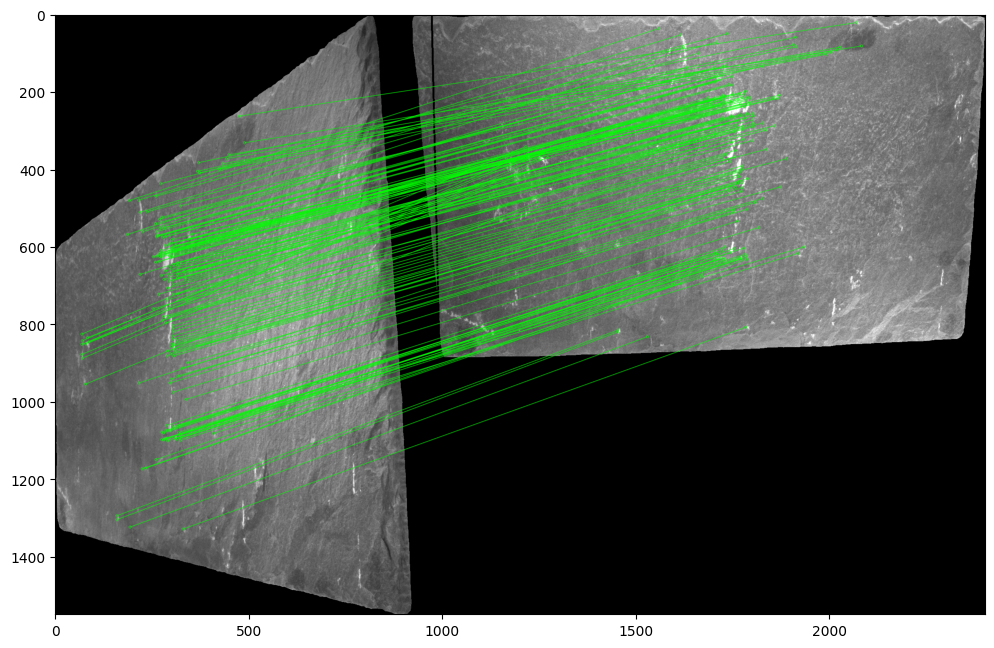}\hfill

\caption{\textbf{Keypoint Matching}. This figure shows the matching of keypoints using the XFeat + LightGlue branch of the hybrid model. The slate tile class is SIN 710, and uses the side and top-down view segmented images.}
\label{fig:1}
\end{figure}

\begin{figure*}[t]
\centering

\includegraphics[width=0.7\linewidth]{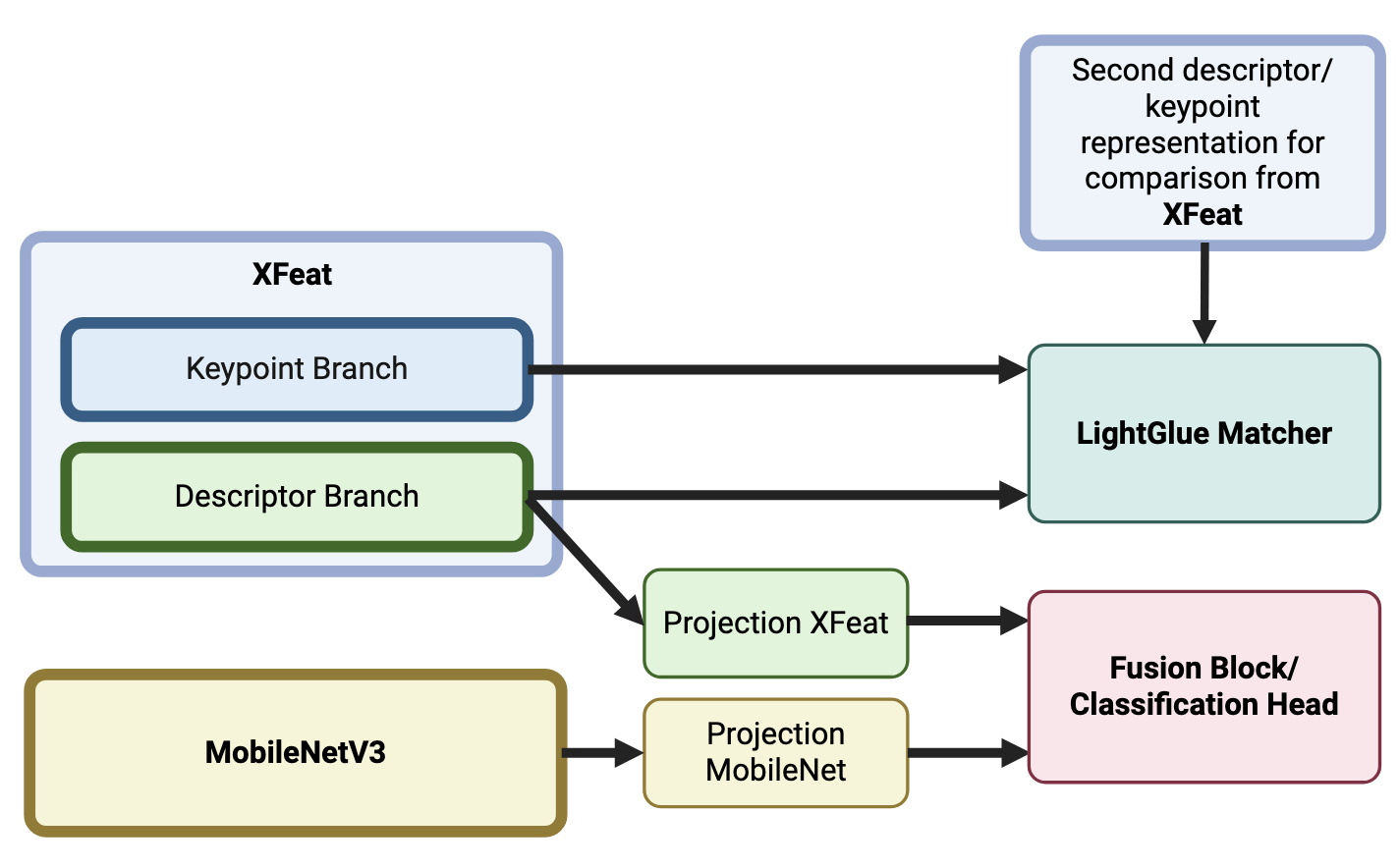}\hfill

\caption{\textbf{Model Structure}. The model has matching and classification parts from XFeat \cite{potjeXFeatAcceleratedFeatures2024} and MobileNetV3 \cite{howardSearchingMobileNetV32019a}. The descriptor branch of the XFeat model feeds into a projection head that is fused with the projection head of the MobileNetV3. This branch is used for classification. On the other hand, the matching branch compares the extracted keypoints and descriptors of two images to find correspondences.}
\label{fig:2}
\end{figure*}

We propose a hybrid, lightweight model, that combines XFeat \cite{potjeXFeatAcceleratedFeatures2024} + LightGlue \cite{lindenbergerLightGlueLocalFeature2023} for matching, with MobileNetV3 \cite{howardSearchingMobileNetV32019a} for classification. The image matching branch is used for re-identifying individual tile instances, while the classification branch is employed to classify the extraction site origins. These are evaluated on the standard MegaDepth-1500 benchmark \cite{liMegaDepthLearningSingleView2018} and a new slate dataset of 2,610 paired top-down and side-view images. The slate images were taken in collaboration with one of Europe's largest slate production and distribution companies: Rathscheck Schiefer und Dachsysteme.

The main contributions of this work are:
\begin{itemize}
\item Experimental design pipeline for generating the slate tile dataset (involving image segmentation).
\item A hybrid architecture combining XFeat + LightGlue with MobileNetV3 for robust slate tile matching and classification.

\item Extensive evolutions involving comparisons on edge devices.

\end{itemize}

\section{Model Structure}
To address the challenges of classification and instance-level re-identification in slate tiles, we fuse XFeat and MobileNetV3 within one framework.

\subsection{\textbf{XFeat}}
XFeat is a state-of-the-art, lightweight convolutional neural network (CNN) \cite{lecunGradientbasedLearningApplied1998} designed for robust keypoint detection and local feature extraction. Unlike traditional detectors (e.g. SIFT) that use fixed filters, XFeat learns to detect keypoints and compute descriptors end-to-end. The model aims to balance speed and accuracy, achieving matching accuracy comparable to larger models such as SuperPoint and DISK, while being up to 5x faster than other lightweight models \cite{potjeXFeatAcceleratedFeatures2024}. XFeat is able to perform sparse matching based on keypoints, as well as dense matching based on its feature map. 

XFeat is trained on a mix of MegaDepth and synthetically warped COCO images with images resized to (W = 800, H = 600) \cite{linMicrosoftCOCOCommon2015}. It outputs three maps for an input image: a 64‑dimensional dense descriptor map \textbf{F}, a keypoint heatmap \textbf{K}, and a reliability/confidence map \textbf{R}. These can be organized into two branches:
\begin{itemize}
\item \textit{Keypoint Head}: the input is partitioned into 8×8 cells, with each cell reshaped into a 64‑dimensional vector. Following this, four 1x1 convolutions are used for regressing keypoint coordinates, creating the embedding $\textbf{K} \in \mathbb{R}^{H/8 \times W/8 \times (64 + 1)}$.

\item \textit{Descriptor Head}: the network uses a feature pyramid strategy, merges multi-scale features, and uses shallow early layers followed by deeper layers to balance speed and robustness. In particular, XFeat merges features from three scales (1/8, 1/16, 1/32 of input) via upsampling and summation, then fuses them to form the final descriptor field $\textbf{F} \in \mathbb{R}^{H/8 \times W/8 \times 64}$. An additional convolutional block is used to regress a reliability map \textbf{R}. This design keeps spatial detail (important for accurate localization) while keeping the compute low \cite{potjeXFeatAcceleratedFeatures2024}.

\end{itemize}

For the matching branch of our pipeline, we employ sparse local features extracted from the dense feature map \textbf{F}, together with keypoints detected by XFeat. In contrast, the classification branch directly utilizes the dense descriptor feature maps \textbf{F} as part of the backbone representation. We hypothesize, XFeat is particularly effective for natural material analysis because its receptive fields are more sensitive to the subtle surface characteristics (e.g., striations and other morphological patterns) in the slate stone.

\subsection{\textbf{LightGlue Matcher}}

We integrate a LightGlue \cite{lindenbergerLightGlueLocalFeature2023} head atop the XFeat backbone, which takes two sets of local features and keypoints, to produce correspondences between them. LightGlue is an efficient evolution of the SuperGlue \cite{sarlinSuperGlueLearningFeature2020}\cite{detoneSuperPointSelfSupervisedInterest2018} architecture, with stacks of transformer-like layers, each followed by a confidence classifier. It uses alternating layers of self-attention (aggregating context within the same image for descriptors and keypoints) and cross-attention (communicating information between the two images being matched) to augment the visual descriptors \cite{lindenbergerLightGlueLocalFeature2023}.

LightGlue distinguishes itself through its adaptive computational mechanism as well:
\begin{itemize}
\item \textit{Early Stopping}: The model iteratively predicts correspondence confidence at each layer. If the confidence surpasses a learned threshold, inference is halted early, preventing redundant computation.

\item \textit{Point Pruning}: The network dynamically discards non-matchable points (outliers or occluded regions) in deeper layers, narrowing the search space to relevant features only.
\end{itemize}

As a result, LightGlue offers efficient, and robust image matching, improving upon the original learned matcher for XFeat, with a small compute trade-off. This is important for our application, where the difficulty of matching slate tiles can vary depending on the viewpoint change (top-down vs. side-view) and surface texture distinctiveness.
\begin{table}[t]
\centering
\caption{\textbf{LightGlue configuration}. A lighter version of LightGlue, with less layers and heads. We use a matcher with 6 layers and 1 head.}
\label{tab:I}

\begin{tabular}{c c c}
\toprule
Method & Layers & Heads \\
\midrule
LightGlue (original) & 9 & 4 \\
LightGlue (ours)     & 6 & 1 \\
\bottomrule
\end{tabular}
\end{table}
\subsection{\textbf{MobileNetV3}}

In addition to XFeat, we employ MobileNetV3 (small) as a classification backbone. MobileNetV3  is a state-of-the-art, efficient feature extractor optimized via hardware-aware Neural Architecture Search (NAS) by the NetAdapt algorithm \cite{howardSearchingMobileNetV32019a}. This is designed to extract high-level semantic representations with minimal latency. MobileNetV3 is pre-trained on the ImageNet dataset \cite{heDeepResidualLearning2015}  \cite{krizhevskyImageNetClassificationDeep2017}.

The architecture builds upon multiple inverted residual with bottleneck blocks introduced in MobileNetV2 \cite{sandlerMobileNetV2InvertedResiduals2019}\cite{howardMobileNetsEfficientConvolutional2017}. Each block starts with a narrow input, expands it via a 1×1 convolution to high-dimensional intermediate layers, then eventually projects back to a narrow output with a linear 1×1 conv. This structure allows the network to perform nonlinear transformations in a high-dimensional space while keeping input and output dimensions small, thereby reducing computational cost \cite{howardSearchingMobileNetV32019a}. Additionally, residual (skip) connections encourage efficient feature reuse and facilitate gradient flow.

MobileNetV3 implements Squeeze-and-Excitation (SE) modules within its bottleneck blocks, an approach previously explored in MnasNet \cite{tanMnasNetPlatformAwareNeural2019}. SE modules act as lightweight attention modules for channels and are implemented after the depthwise convolution within the expansion phase. This improves accuracy, as it is performed on the largest representation. 

A further improvement of the MobileNetV3, is that the HardSwish function is applied throughout the model (in addition to ReLU). This is a modified activation function that improves accuracy while being computationally cheaper to calculate than standard Swish \cite{howardSearchingMobileNetV32019a}\cite{howardSearchingMobileNetV32019a}. For an input $x$, HardSwish sets the function to 0 if $x \leq -3$, sets it to $x$ if $x \geq +3$, and to $x \cdot (x + 3)/6$ otherwise.

\subsection{\textbf{Hybrid Model}}

The proposed hybrid model consists of two branches (Fig. \ref{fig:2}): a classification branch and an image matching branch. The image matching branch leverages local descriptors extracted from the dense XFeat feature map \textbf{F} along with keypoints \textbf{K} as input to a LightGlue matcher. We implement a modified, smaller LightGlue head from the XFeat repository, consisting of only six layers and a single attention head (illustrated in Table \ref{tab:I}). The model weights were trained using GlueFactory, a framework designed for training and evaluating deep neural networks such as LightGlue and GlueStick, which specialize in extracting and matching local visual features \cite{lindenbergerLightGlueLocalFeature2023}\cite{pautratGlueStickRobustImage2023}. 
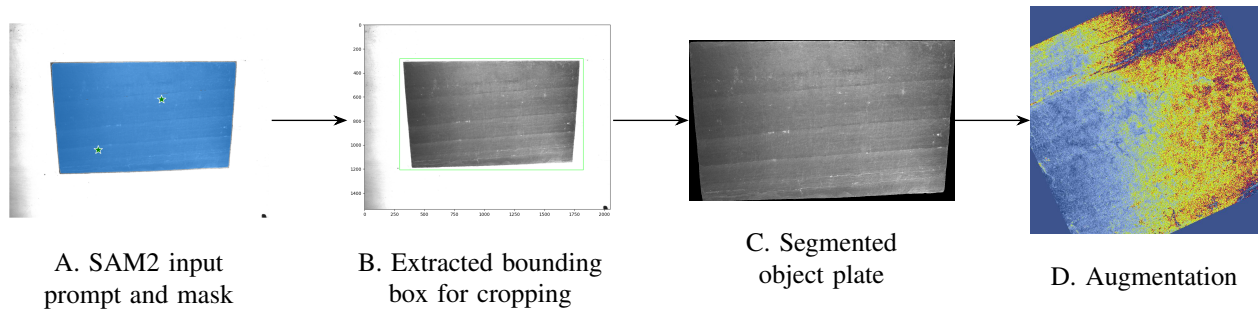
\begin{figure*}[t]
\centering
\begin{tikzpicture}[
    node distance=1cm,               
    every node/.style={align=center},
    img/.style={inner sep=0pt},
    arrow/.style={->, line width=0.5pt, >=Stealth}
]

\node[img] (i1) {\includegraphics[width=3.5cm]{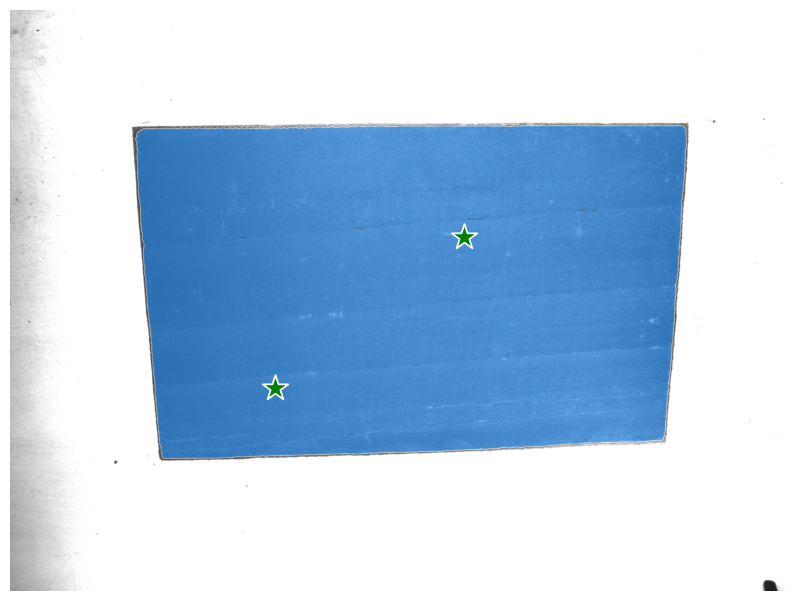}};
\node[img, right=of i1] (i2) {\includegraphics[width=3.5cm]{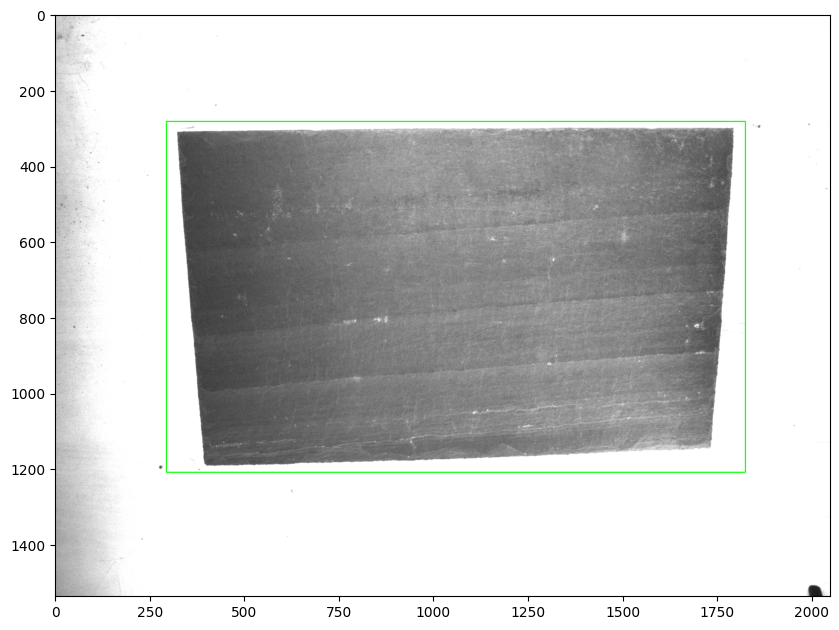}};
\node[img, right=of i2] (i3) {\includegraphics[width=3.5cm]{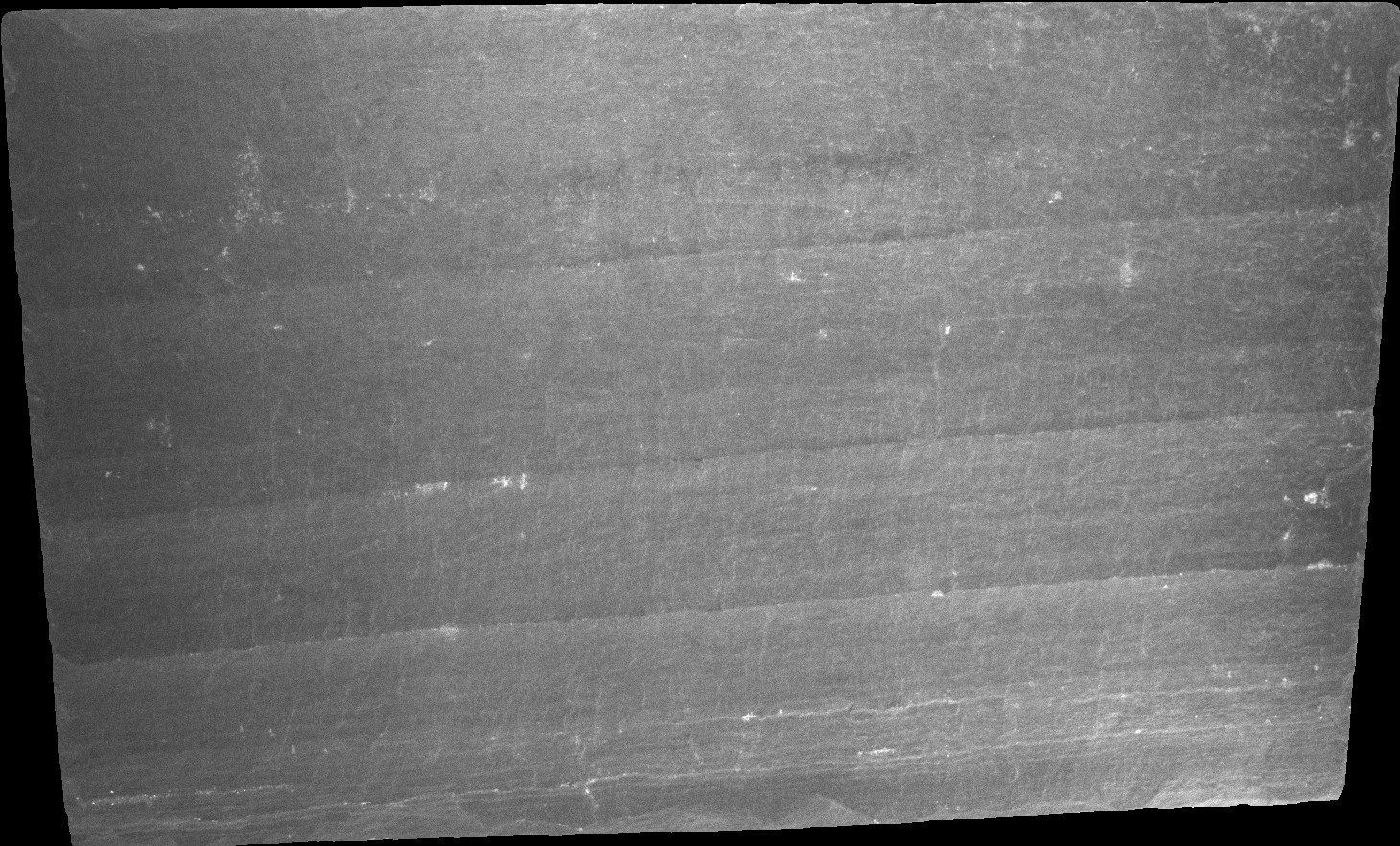}};
\node[img, right=of i3] (i4) {\includegraphics[width=3cm]{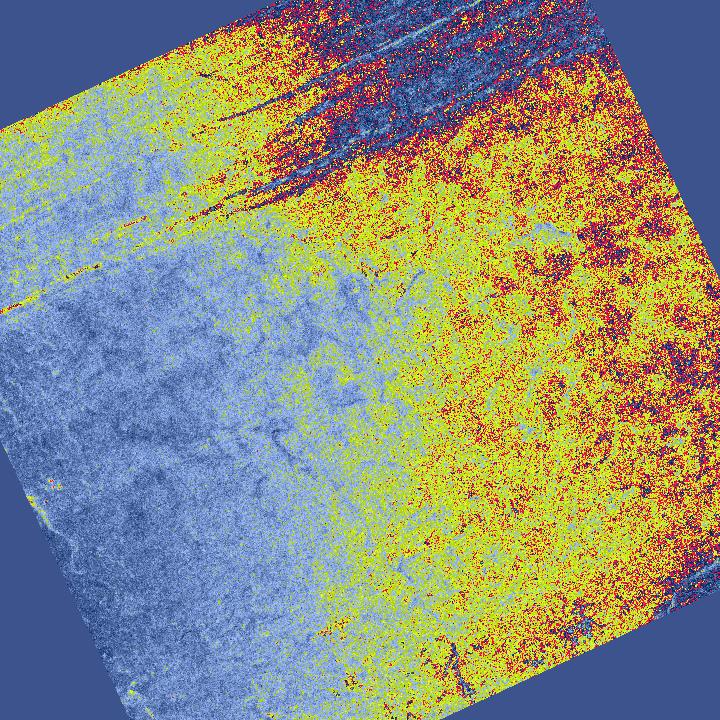}};

\node[below=3mm of i1, text width=3.5cm, align=center] {A. SAM2 input prompt and mask};
\node[below=3mm of i2, text width=3.5cm, align=center] {B. Extracted bounding box for cropping};
\node[below=3mm of i3, text width=3.5cm, align=center] {C. Segmented object plate};
\node[below=3mm of i4, text width=3.5cm, align=center] {D. Augmentation};

\draw[arrow] (i1) -- (i2);
\draw[arrow] (i2) -- (i3);
\draw[arrow] (i3) -- (i4);

\end{tikzpicture}
\caption{\textbf{Dataset generation pipeline.} The original image is segmented using SAM2, using two points as input. The generated mask is used to create the segmented image and the generated bounding box is used to crop. During training this image is cropped again to 720x720 px and augmented.}
\label{fig:5}
\end{figure*}
For classification, the penultimate MobileNetV3 embeddings \textbf{S} (576-dimensional) are projected and fused with the output features from the XFeat projection head. XFeat descriptor output is lower-dimensional than the MobileNetV3 embeddings (64 vs. 576), and both feature sets are projected to a common dimensionality (384, 448, or 512). Each projection head consists of a linear layer, LayerNorm, and a non-linear activation function (ReLU or HardSwish).

The dense descriptor map \textbf{F} from XFeat is fed into its projection head, while the MobileNetV3 backbone outputs features into a separate projection head with the same structure. The resulting embeddings are concatenated and passed through a two-layer fusion block before producing the final classification logits. In the fusion block, the first layer reduces the feature dimensionality by a factor of two, followed by the activation function, dropout, and a second linear layer that outputs the class logits. Prior to the projection heads, the outputs of both backbones are average pooled to reduce spatial dimensions and stabilize feature representations.

\section{Dataset Setup and Methods}
\subsection{\textbf{Dataset Acquisition and Hardware}}

The dataset constructed for this study consists of 2,610 images of slate tiles sourced from six distinct Spanish deposits (designated as classes: SIN 120, SIN 150, SIN 320, SIN 340, SIN 360, and SIN 710). The physical dimensions of the slate plates are standardized at 40×25 cm. 

Image acquisition was performed using two monochrome Imaging Source cameras (Model DMK 37AUX252) equipped with a 6mm lens, and a resolution of 2048×1536 pixels. To ensure consistent illumination and minimize reflections, the scene was lit by an overhead lighting source (Makita 18V LED).

Data collection involved capturing each slate plate from two distinct angles: a top-down view and a side view. This resulted in 1,305 images per view angle. The dataset is used differently depending on the task under consideration:
\begin{itemize}
\item \textit{Classification Task}: Utilizes exclusively the top-down view images.

\item \textit{Image Matching Task}: Utilizes both top-down and side-view images to evaluate alternate view matching performance.
\end{itemize}

\subsection{\textbf{Dataset Characteristics and Ambiguity}}
In practice, differentiation between slate quarries relies on a combination of visual cues, such as surface texture and color, as well as acoustic and physical characteristics. A key challenge in this dataset arises from the strong visual and geological similarity between certain classes. For instance, the SIN 150 and SIN 360 classes originate from adjacent quarries and are therefore geologically identical, making visual discrimination difficult even for domain experts when relying solely on images. SIN 320 and SIN 360 are visually similar as well. In contrast, classes such as SIN 340 exhibit distinct morphological characteristics, such as a rustic surface texture and horizontal striations, making them distinguishable even to non-experts.

We benchmarked the proposed deep learning model against a human expert from Rathscheck with over 30 years of professional experience in slate classification. The expert is capable of confidently distinguishing among a wide range of slate classes. For this comparison, the held-out validation set (259 images) was randomly sampled, and the expert was asked to assign a class label to each image based solely on visual inspection. The expert estimated that image-only classification would yield an accuracy of approximately 70\%; or 85\% if provided with a physical slate sample (allowing better access to visual, acoustic, and physical cues).

\subsection{\textbf{Preprocessing and Segmentation}}

The slate plate images are segmented from the background to isolate the Region of Interest (ROI). We employ the Segment Anything Model 2 (SAM2) \cite{raviSAM2Segment2024} for this purpose, utilizing a two-point input prompting strategy to generate precise masks for the slate tiles. The resulting masks and bounding boxes are then used to tightly crop the images around the tile boundaries and place the extracted pixel data onto a black background. Images of insufficient quality or with incomplete segmentation due to interference are removed from the dataset.
\begin{table}[t]
\centering
\caption{\textbf{Dataset class distribution.}}
\label{tab:II}

\begin{tabular}{c c c}
\toprule
Class & Train & Validation \\
\midrule
SIN 120 & 201 & 50\\
SIN 150 & 201 & 50\\
SIN 320 & 163 & 40\\
SIN 340 & 159 & 39\\
SIN 360 & 161 & 40\\
SIN 710 & 161 & 40\\

\bottomrule
\end{tabular}
\end{table}

\subsection{\textbf{Experimental Configuration}}
\subsubsection{Classification Dataset}
The top-view dataset is partitioned into a training set of 1,046 images and a validation set of 259 images. This split is balanced class-wise to ensure equal representation across all six classes as outlined in Table \ref{tab:II}.

We adopt a 5-fold cross-validation strategy on the original training set, using an 80/20 split for each fold (yielding 837 training and 209 validation images per fold). The model with the highest accuracy on the internal validation split is selected for final evaluation on the held-out validation set.
\begin{table*}[b]
\centering
\caption{\textbf{Human Expert Per Class Recall}. HardSwish 512 dimension projection head comparison to expert.}
\label{tab:III}

\begin{tabular}{c c c c c c c}
\toprule
Method & SIN 120 & SIN 150 & SIN 320 & SIN 340 & SIN 360 & SIN 710 \\
\midrule
Human Expert & \textbf{100} & 100 & 53.8 & \textbf{100} & 38.4 & 98\\
\textbf{Hybrid}     & 98 & \textbf{100} & \textbf{87.5} & 97.4 & \textbf{87.5} & \textbf{100}\\

\bottomrule
\end{tabular}
\end{table*}

To improve model generalization and mitigate overfitting, we apply a robust data augmentation pipeline during training. This includes:

\begin{table*}[t]
\centering
\caption{\textbf{Projection Head and Activation Function Adjustments}. Validation accuracy is calculated on the held-out validation set, using the best model out of all folds. The Mean Fold Accuracy is the average validation accuracy of the split (per model across all 5 folds). }
\label{tab:IV}

\begin{tabular}{c c c c c c c c}
\toprule
Model & Projection Head & Activation Function & Total Params & Trainable Params & Val Accuracy & F1 & Fold Accuracy\\
\midrule
Hybrid & 384 & Hardswish & 3.0M & 545K & 94.2 & 93.8&94.6\\
Hybrid     & 448 & Hardswish & 3.2M & 693K& 95.3 & 95.0 & 95.0\\
Hybrid     & 512 & Hardswish & 3.3M & 858K&\textbf{95.4} & \textbf{95.1}& \textbf{95.4}\\

Hybrid  & 384 & ReLU & 3.0M & 545K& 95.3&95.0&95.5\\
Hybrid     & 448 & ReLU &  3.2M & 693K& 94.6&94.3&94.1\\
Hybrid     & 512 & ReLU & 3.3M & 858K&94.2&93.8&\textbf{95.6}\\
\midrule
MobileNetV3, 1 layer & - & - & 930K & 3.5K & 84.5 & 83.9&89.0\\

MobileNetV3, 2 layer & - & Hardswish & 1.3M & 335K& 91.1&90.1&93.6\\

MobileNetV3, 2, expanded & - & Hardswish & 1.6M & 671K& 94.2&93.7&93.2\\

MobileNetV3, 2, expanded & - & ReLU & 1.6M & 671K& 93.8&93.3&93.5\\
MobileNetV3, 3 layer & - & Hardswish & 1.6M & 668K& 91.9&91.7&93.1\\
MobileNetV3, 3 layer & - & ReLU & 1.6M & 668K& 93.8&93.5&93.0\\
\bottomrule
\end{tabular}

\end{table*}
\begin{itemize}
\item \textit{Random Cropping}: Some classes are cut differently than others, which is noticeable when observing the edges of the slate (e.g., white edge lines visible in certain classes). To mitigate these biases, we randomly extract 720 × 720 pixel patches from the source images. This strategy suppresses edge-related cues and encourages the model to focus on intrinsic surface texture and structural characteristics of the slate material. 

\item \textit{Geometric Transformations}: Random horizontal flips, rotations, and affine transformations.

\item \textit{Photometric Distortions}: Color jitter (brightness and contrast adjustments).
\end{itemize}

The model uses layer normalization and dropout as further regularization techniques.

\subsubsection{Matching Dataset}
To quantify image matching performance, we evaluate our model on the standard MegaDepth-1500 benchmark \cite{liMegaDepthLearningSingleView2018}, a widely used dataset for evaluating local feature matching and relative pose estimation. MegaDepth-1500 consists of challenging image pairs with viewpoint and illumination changes, allowing for robust assessment of image matching quality. We also assess matching accuracy on our novel slate dataset using images from both camera viewpoints (2,610 images in total).

\section{Experiments and Results}
\subsection{\textbf{Classification}}
\subsubsection{\textbf{Setup}}
We train the hybrid model, consisting of MobileNetV3-small and XFeat backbones, in a supervised manner on the top-down slate dataset using 5-fold cross-validation (1,046 images) for 100 epochs. In addition, a MobileNetV3-small model is trained as a baseline. All models are implemented in PyTorch and trained on an NVIDIA GPU Quadro RTX 5000. The models are evaluated on a CPU Intel Core i7-9800X @ 3.80GHz x 16 and a Raspberry Pi 5 for edge comparison. The AdamW optimizer is used with an initial learning rate of 0.001, momentum of 0.9, and weight decay of 0.999. The batch size is set to 16, and input images are randomly cropped to a resolution of 
720×720 pixels to preserve surface detail. We employ a cross-entropy loss function, a dropout rate of 0.2, and data augmentation as described in the previous section.

Both backbone networks are initialized with pretrained weights and remain frozen during training to prevent overfitting; only the parameters of the projection and fusion blocks (i.e., the classification head) are optimized. We investigate multiple classification head configurations, as outlined in Table \ref{tab:IV}. This involves adjusting the dimensional output of the projection head to 384, 448 or 512 channels. Additionally, we study the impact of activation functions within the classification head, comparing ReLU and HardSwish, as HardSwish may better preserve the representational properties of MobileNet features.

Our design hypothesis is that, as a matcher and keypoint detector, XFeat produces representations that emphasize the fine-grained surface characteristics of the slate material. When combined with the more global semantic features extracted by MobileNetV3, this hybrid representation is expected to improve discriminative performance, particularly for visually ambiguous classes.

\subsubsection{\textbf{Evaluation}}

We compare the proposed hybrid architecture against a simple MobileNetV3-small baseline with a single linear classification layer. As with the hybrid model, the backbone is frozen and only parameters in the classification head are updated during training. Additionally, we investigate MobileNetV3 variants with larger classification heads, allowing a more controlled assessment of whether performance gains stem from architectural complementarity rather than increased model capacity alone. The larger classification head consists of two or three linear layers. The first layer optionally expands the input dimensionality by a factor of two, followed by a non-linear activation function. This is followed by one or more additional linear layers. The final layer projects the resulting features to the target number of classes. 

Standard classification metrics (Accuracy, Precision, Recall, F1 Score) are used to quantify the performance of both models. 

\subsubsection{\textbf{Results}}
Across all models, fold 4 achieved the highest accuracy, with the hybrid models (projection dim 384, 448, 512, ReLU activation function; projection dim 384, 448, 512, HardSwish activation function) approaching 98\% accuracy. The MobileNet baselines achieved accuracy around 96\% for fold 4. However, it is important to note that this result may represent a favorable sampling effect, as the remaining folds exhibited accuracies in the range of 90–95\%. We therefore look at the average validation accuracy across all folds as well, and notice that all the hybrid models outperform the MobileNets in this regard. Overall, the best-performing configuration was the hybrid model with a projection dimension of 512 and HardSwish activation function (Table \ref{tab:IV}). 

The 512 HardSwish model achieved 95.4\% accuracy on the held-out validation set, a +10.9\% improvement compared to the MobileNet single-layer model. For the MobileNet baseline, the best performing model was the two-layer expanded model using the HardSwish activation function in the classification head, with a validation accuracy of 94.2\%. In conclusion, the best hybrid model outperforms all baseline MobileNetV3 configurations, and exhibits higher mean fold accuracy as well.

For comparison between the performance of the human expert and the classification model, we choose the best performing model in regards to held-out validation accuracy and mean fold accuracy (512, HardSwish). Comparison between the human expert and the hybrid deep learning model exhibits similar trends across classes. The model classifies classes such as SIN 120, SIN 150, SIN 340, and SIN 710 with high recall, close to or better than the human expert. However, the expert and the model showed lower performance for visually ambiguous classes, such as SIN 320 and SIN 360. Notably, the hybrid model outperformed the expert on these challenging classes, achieving 87.5\% versus 53.8\% for SIN 320 and 87.5\% versus 38.4\% for SIN 360, a substantial increase (Table \ref{tab:III}). 

It is important to consider that the human expert typically relies on additional cues, such as physical and acoustic information, for deposit identification, and the captured images may differ from the visual conditions the expert is accustomed to, including variations in lighting.

\subsection{\textbf{Image Matching}}
\subsubsection{\textbf{Setup}}
The XFeat + LightGlue branch of our proposed hybrid model combines XFeat as a local descriptor and LightGlue as a learned matcher. For this branch, the maximum number of keypoints detected per image is set to 4,096, and the maximal image dimension is constrained to 1,024 pixels. We compare the hybrid model against the original XFeat and XFeat*, where XFeat* is an enhanced version of XFeat that uses 10,000 local features for semi-dense matching and improving correspondence robustness. Weights for the XFeat backbone and LightGlue matcher are selected from the XFeat repository: \href{https://github.com/verlab/accelerated_features}{github.com/verlab/accelerated\_features}. 
\subsubsection{\textbf{Evaluation}}
We evaluate the hybrid model on the MegaDepth-1500 benchmark for relative pose estimation, computing the AUC at thresholds 5, 10, and 20\degree. An LO-RANSAC threshold of 2.5 is chosen for matching. This is performed using modified evaluation scripts from the XFeat repository, to enable evaluation on LightGlue matches \cite{potjeXFeatAcceleratedFeatures2024}.

Furthermore, we evaluate the matching accuracy of the hybrid model applied to our slate dataset of 2,610 images. Full-size crops are used for evaluation, resulting in image dimensions of approximately 1,480 × 890 px for top-down views and 950 × 1,530 px for side views. For each slate tile, exactly one image is available from each viewpoint, enabling assessment of the model’s ability to match the same physical object across perspective changes.

Matching performance is evaluated using top-1 and top-5 accuracy. Images are ranked based on the number of feature correspondences produced by the matcher, and a match is considered correct if the paired image from the alternate viewpoint is retrieved as the top-ranked candidate (top-1) or appears within the top five candidates (top-5).

\begin{figure}[t]
\centering
\begin{tikzpicture}[
    node distance=0.8cm, 
    every node/.style={align=center},
    img/.style={inner sep=0pt}
]

\node[img] (i1) {\includegraphics[width=0.4\textwidth]{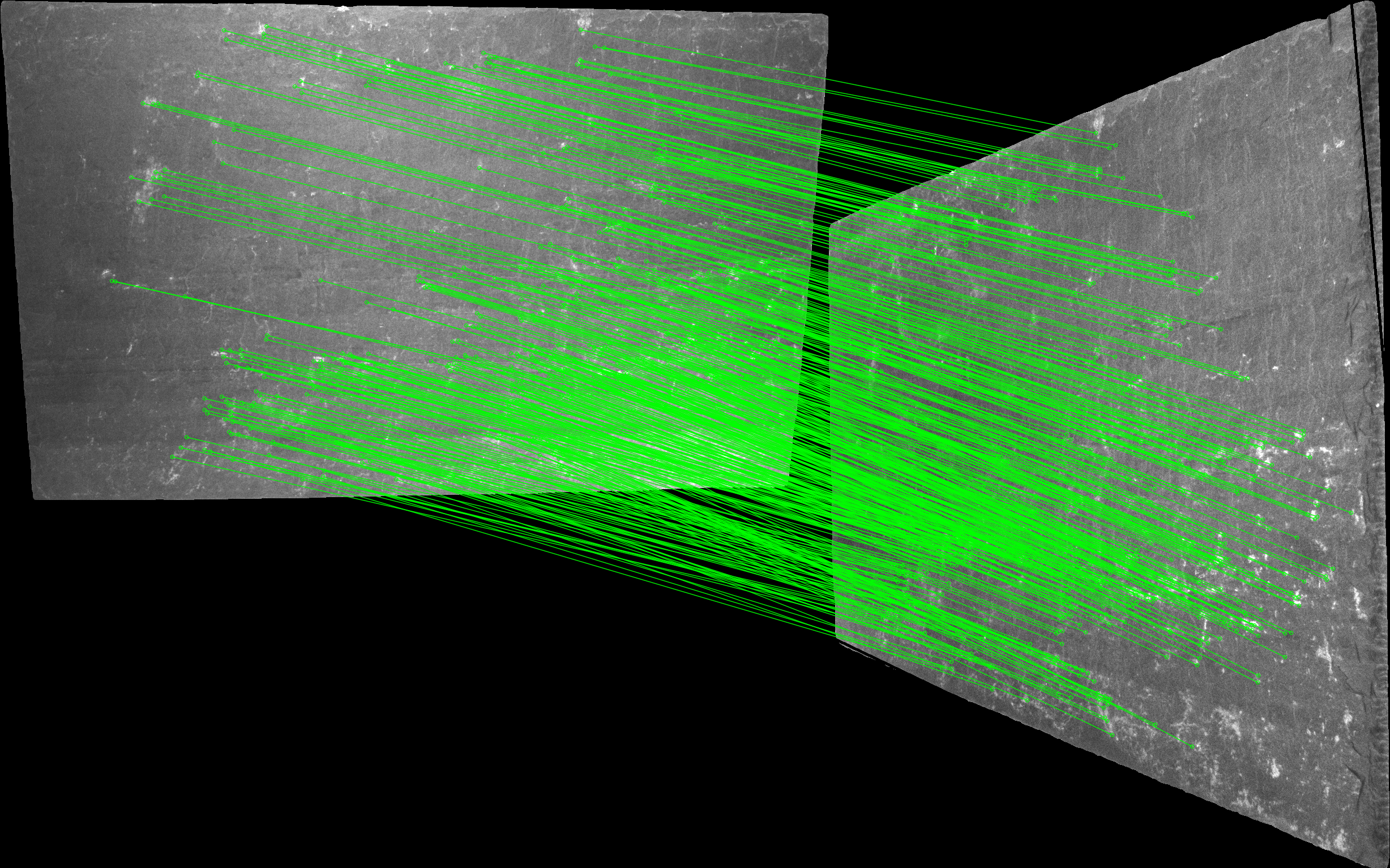}};
\node[img, below=of i1] (i2) {\includegraphics[width=0.4\textwidth]{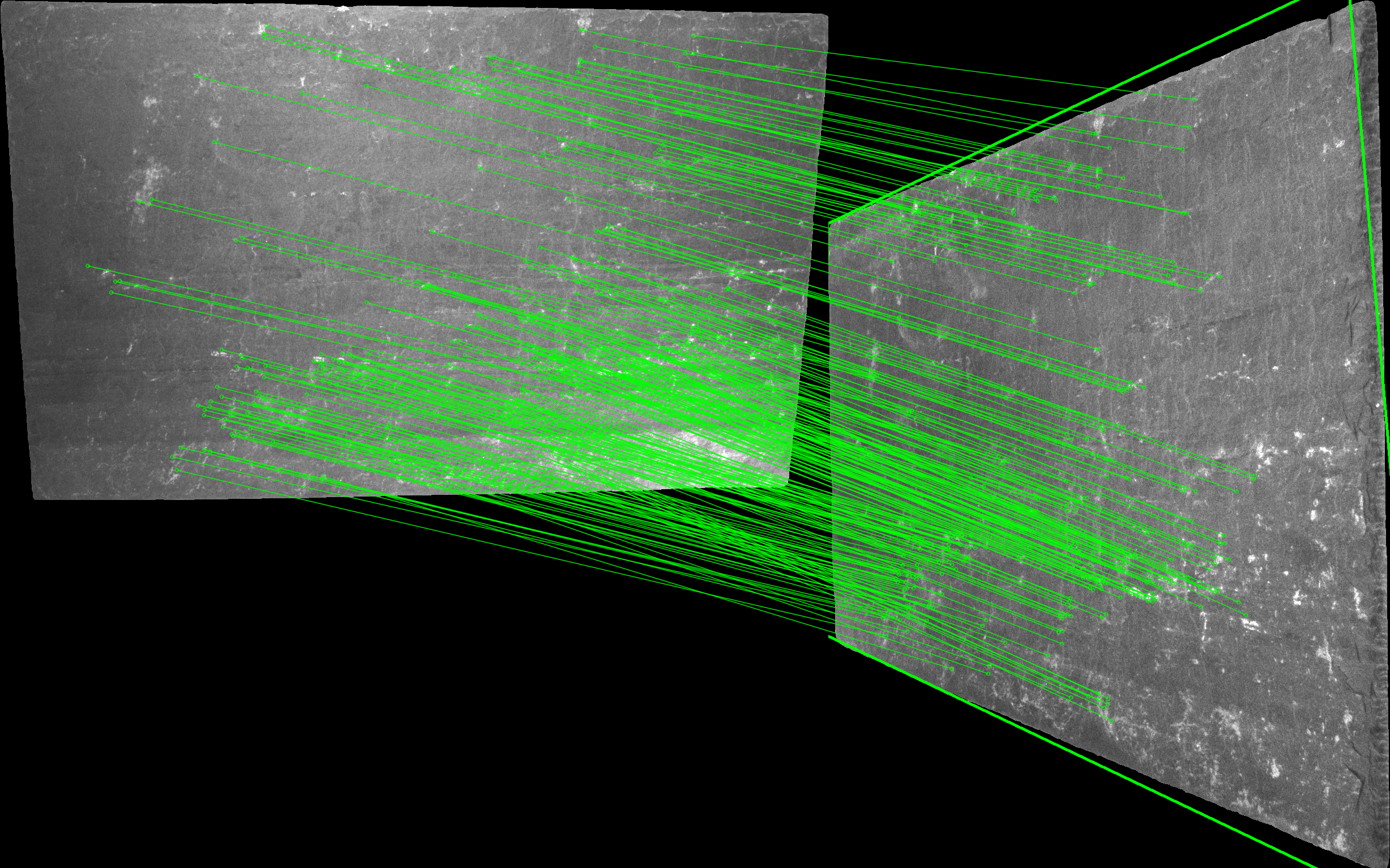}};

\node[below=2mm of i1, text width=0.3\textwidth] {(a) LightGlue matching};
\node[below=2mm of i2, text width=0.3\textwidth] {(b) XFeat matching};

\end{tikzpicture}
\caption{\textbf{Matching task.} Visualization of keypoint matching on SIN 340 slate plates, used to calculate the top-k retrieval accuracy. We can see here that LightGlue finds more detailed and correct matches compared to the original XFeat matcher.}
\label{fig:three_images_tikz}
\end{figure}
\subsubsection{\textbf{Results}}
The hybrid XFeat + LightGlue model shows substantial improvement over the baseline models. Compared to the original XFeat, the hybrid model achieves AUC gains of +14.4, +15.4, and +14.8 at the 5\degree, 10\degree, and 20\degree, thresholds respectively. Compared to the semi-dense XFeat* matcher, it provides additional AUC gains of +6.8, +6.4, and +5.4, as summarized in Table \ref{tab:V}. 
\begin{table}[t]
\centering
\caption{\textbf{Relative Pose Estimation on MegaDepth-1500.} XFeat performance as reported in its paper. The LightGlue head is trained to match maximum 4096 keypoints}
\label{tab:V}

\begin{tabular}{c c c c}
\toprule
Method & AUC @5\degree & AUC @10\degree & AUC @20\degree \\
\midrule
XFeat (original) & 42.6 & 56.4 & 67.7\\
XFeat*     & 50.2 & 65.4 & 77.1 \\
\textbf{XFeat + LightGlue (4096)}     & \textbf{57.0} & \textbf{71.8} & \textbf{82.5}\\

\bottomrule
\end{tabular}

\end{table}
On the slate dataset, the hybrid model achieves a matching accuracy of 91.7\% with a top-k of 5 in comparison to just an XFeat backbone (89\%), demonstrating its robustness in handling real-world viewpoint changes. The top-1 retrieval accuracy is also higher in the LightGlue model compared to the original XFeat matcher as illustrated in Table \ref{tab:VI}.

\begin{table}[H]
\centering
\caption{\textbf{Matching Performance on Slate Dataset.} Top-k retrieval accuracy based on top-down and side-view matching.}
\label{tab:VI}

\begin{tabular}{c c}
\toprule
Method & Accuracy \\
\midrule
XFeat (original) top-1 & 87.6\\
XFeat (original) top-5 & 89.2\\
\textbf{XFeat + LightGlue top-1}     & \textbf{90.0}\\
\textbf{XFeat + LightGlue top-5}     & \textbf{91.7}\\

\bottomrule
\end{tabular}

\end{table}

\subsection{\textbf{Feature extraction speed}}
To evaluate the speed of feature extraction with respect to edge applicability, we benchmarked our method on a standard CPU and a Raspberry Pi 5. For the classification branch, we observed only a small increase in inference time compared to the pure MobileNet baseline. However, for the matching branch, comparing XFeat-only matching with XFeat + LightGlue, the processing time per image increased by approximately one order of magnitude. For example, on an Intel i7 CPU, the complete LightGlue-based matching pipeline required around 1058 ms per image, whereas the XFeat-only pipeline took approximately 126 ms. While the classification branch is able to process images in as little as 35.6 ms per image, future improvements to the matching branch may include further pruning of the matching head to reduce latency. All reported results correspond to average inference times computed over 100 images of sizes 640x480 and 1480x890.

\begin{table}[H]
\centering
\caption{\textbf{Feature extraction.} Speed of the classification branch on a Raspberry Pi 5 (RP5) and Intel Core i7 CPU.}
\label{tab:VII}

\begin{tabular}{c c c c}
\toprule
Method & Device &Time per image (ms) & Image Size \\
\midrule
MobileNet (original)  & CPU & 19.5 &640x480\\
MobileNet (original)  & CPU & 64.9 &1480x890\\
MobileNet (original)  & RP5 & 64.7 &640x480\\
MobileNet (original)  & RP5 & 661.8 &1480x890\\
Hybrid     &  CPU & 35.6&640x480 \\
Hybrid     & CPU & 135.6&1480x890 \\
Hybrid     &  RP5 & 94.4&640x480 \\
Hybrid     & RP5 & 1359.3&1480x890 \\
\bottomrule
\end{tabular}

\end{table}

\section{Conclusion and Further Steps}
In this work, we presented a hybrid deep learning model that jointly addresses instance-level image matching and slate tile classification within a unified architecture. By combining XFeat-based dense descriptors with a compact LightGlue matching head and MobileNetV3-based semantic embeddings, the proposed model effectively captures both fine-grained surface details and global contextual information. The integration of LightGlue improves pose estimation and matching performance over the original XFeat and semi-dense XFeat* approaches, while maintaining computational efficiency suitable for practical deployment.

Experimental results on the MegaDepth-1500 benchmark and a newly introduced slate dataset demonstrate the robustness of the proposed model under challenging viewpoint changes. Furthermore, the classification results show near or better than human expert performance with improved accuracy on visually ambiguous classes, highlighting the benefit of feature fusion.

Future work could explore the development of a multimodal model which accepts acoustic and visual information as input, creating a more robust pipeline for classification. Moreover, the matching pipeline is not invariant to rotation, and we suggest incorporating rotation-invariant learned steerers (with simple CNNs) \cite{bokmanSteerersFrameworkRotation2024}. This approach is lightweight and robust, promising an improvement in performance in real-world settings. Finally, extending the model toward a foundation architecture for industrial natural materials represents a promising direction for future work.

\bibliographystyle{ieeetr}

\bibliography{references}

\end{document}